\documentclass{article}
\usepackage{ijcai23}

\usepackage{times}
\usepackage{soul}
\usepackage{url}
\usepackage[hidelinks]{hyperref}
\usepackage[utf8]{inputenc}
\usepackage[small]{caption}
\usepackage{graphicx}
\usepackage{amsmath}
\usepackage{amsthm}
\usepackage{booktabs}
\usepackage{algorithm}
\usepackage{algorithmic}
\usepackage[switch]{lineno}
\usepackage[square,sort&compress,numbers]{natbib}
\usepackage{color}
\usepackage{xcolor}
\usepackage{colortbl}
\definecolor{mygray}{gray}{.9}
\definecolor{mypink}{rgb}{.99,.91,.95}
\definecolor{mycyan}{cmyk}{.3,0,0,0}
\usepackage{stfloats}
\usepackage{multirow}
\title{An Effective Data Creation Pipeline to Generate High-quality Financial Instruction Data for Large Language Model}

\author{
Ziao Wang
\and
Jianning Wang\and
Junda Wu\and
Xiaofeng Zhang\footnote{Corresponding author. Email: zhangxiaofeng@hit.edu.cn}\and
\affiliations
School of Computer Science and Technology, Harbin Institute of Technology, Shenzhen, China\\
\emails
wangziao1993@hotmail.com, jianning.wang@outlook.com, wujunda@stu.hit.edu.cn, zhangxiaofeng@hit.edu.cn
}

\begin{document}

\maketitle

\begin{abstract}
At the beginning era of large language model, it is quite critical to generate a high-quality financial dataset to fine-tune a large language model for financial related tasks. Thus, this paper presents a carefully designed data creation pipeline for this purpose. Particularly, we initiate a dialogue between an AI investor and financial expert using ChatGPT and incorporate the feedback of human financial experts, leading to the refinement of the dataset. This pipeline yielded a robust instruction tuning dataset comprised of 103k multi-turn chats. Extensive experiments have been conducted on this dataset to evaluate the model's performance by adopting an external GPT-4 as the judge. The promising experimental results verify that our approach led to significant advancements in generating accurate, relevant, and financial-style responses from AI models, and thus providing a powerful tool for applications within the financial sector.
\end{abstract}

\section{Introduction}
In recent years, the abilities of pre-trained language models have experienced significant enhancements, with instruction-tuning playing a pivotal role in these advancements. Despite these improvements, a critical gap remains in the application of these models within the financial sector, 
i.e., inaccurate or irrelevant responses from these models could lead to significant financial implications, underscoring the importance of enhancing their performance for financial tasks.

In the literature, several approaches have been proposed \citet{alpaca:rohan:2023,baize:xu:2023} to generate instruction data to fine-tune a large language model on the down-stream task. Although these approaches well employed the SOTA LLMs like GPT-4 \cite{openai2023gpt4}, the quality of the generated data are not suitable for financial related tasks. 
To cope with these challenges, this paper introduces a novel data creation pipeline designed for these high-stakes domains. The cornerstone of our approach is to leverage the in-context learning ability of contemporary large language models \cite{wei:llama:2023,chen:ImprovingInContextFewShot:2022} and supplement it with a high-quality corpus sourced directly from the financial domain, such as financial reports. This allows us to provide both real-world information and processed knowledge, thereby improving the model's ability to generate accurate and relevant responses.

\begin{figure}[!t]
  \centering
  \includegraphics[width=0.99\linewidth]{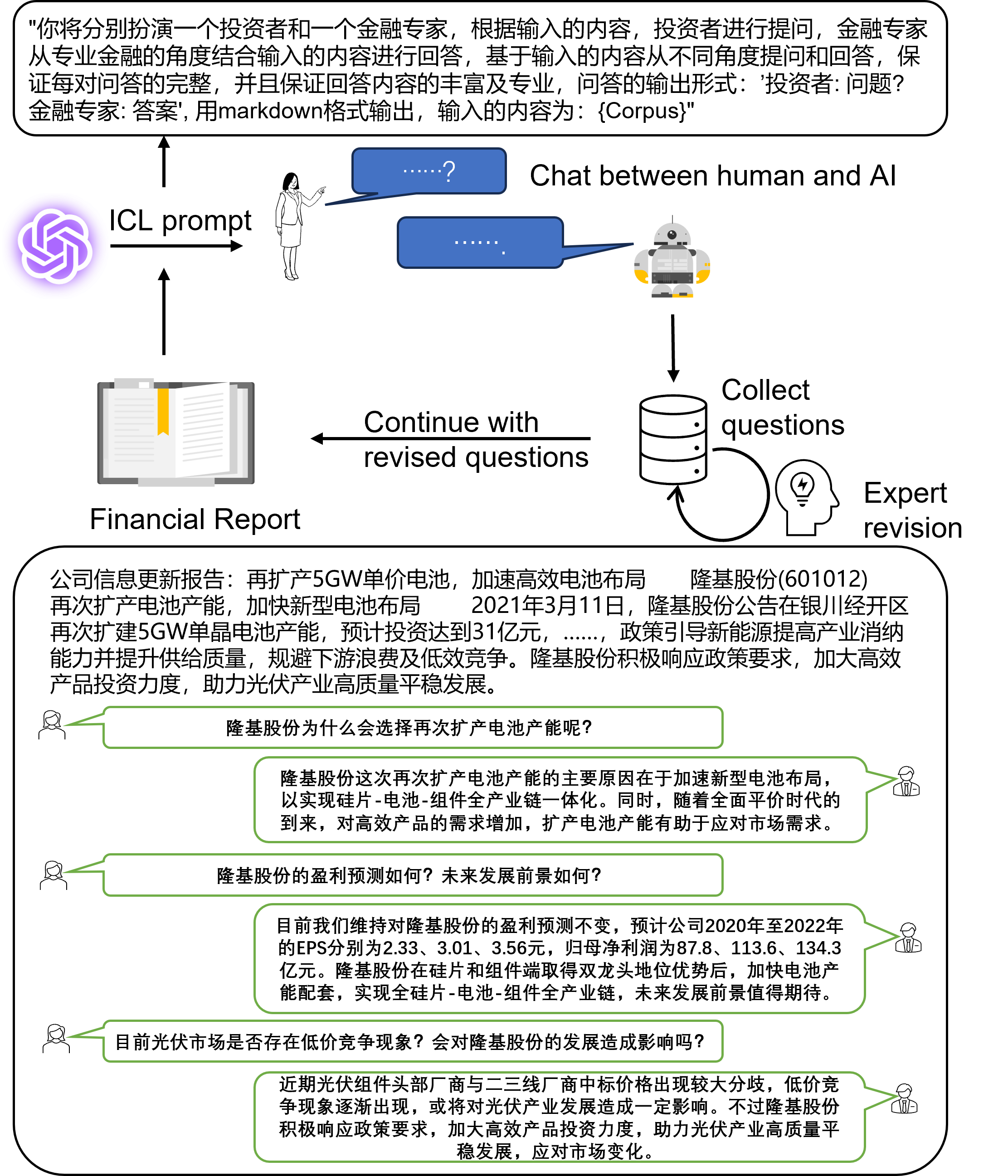}
  \caption{An illustrating example of the data collection pipeline.}\label{fig:example}
\end{figure}

As illustrated in Figure \ref{fig:example}, our approach begins with a high-quality corpus such as a financial report. Using ChatGPT's strong in-context learning ability, we simulate a dialogue between an investor and a financial expert. The generated questions are clustered and then shown to human financial experts for refinement and feedback. This process helps in pruning low-quality or non-financial style questions, which are then sampled to continue the data collection pipeline. Our method resulted in a dataset consisting of 103k multi-turn chats, which was used for instruction tuning on open-sourced language models such as LLama \cite{wei:llama:2023}. We further conducted comprehensive tests on the models' performance on financial tasks, including a GPT-4 evaluation with specifically created questions. The contributions of this paper are summarized as follows:
\begin{itemize}
    \item We design a novel data creation pipeline specifically designed for high-stakes domains, such as finance, and we generate a high-quality instruction tuning dataset consisting of 103k multi-turn chats.
    \item We automatically generate a set of high quality financial questions which could be used to comprehensively evaluate the model performance.
    \item We trained a large language model using this dataset and extensively evaluated the model performance. Both the dataset as well as the questions will be released later for public use. 
\end{itemize}

\section{Related Work}
Instruction tuning has emerged as a fundamental component in the advancement of pre-trained language models. The large language models such as GPT-3 and ChatGPT have achieved substantial enhancements using this technique. The underlying idea is to align the model's behavior with human values in a specific domain by utilizing an instructional dataset. 

\paragraph{Instruction-tuning on LLM}\citet{ouyang:instructgpt:2022} introduced InstructGPT, designed to comply with instructions presented in natural language and produce useful responses. The model performance exceeded that of GPT-3 after tuning under the instruction, substantiating the beneficial impact of instruction tuning. \citet{databricks:dolly:2023} presented Dolly, an instruction-tuned large language model. Dolly was trained on a dataset amalgamated from various sources, including the InstructGPT dataset by OpenAI and a fresh dataset crafted by Databricks. This new dataset encompasses instructions and outputs from real-world tasks undertaken by data scientists and engineers. Dolly exhibited outstanding capabilities in tasks that necessitated understanding and generating intricate text, alongside following instructions. \citet{köpf:openassistant:2023} proposed a large-scale dataset, OpenAssistant Conversations, intended to expedite the advancement of aligned language models. This dataset is the product of a crowdsourcing endeavor and includes over 1.5 million messages from an excess of 10,000 contributors. The authors also showcased the results of a user preference study, which established that a model, Pythia-12B, fine-tuned on this dataset, represents a formidable rival to OpenAI's gpt-3.5-turbo model. \citet{vicuna2023}  is an open-source chatbot that was fine-tuned on user-shared dialogues gathered from ShareGPT by using the LLaMA model \cite{wei:llama:2023} and has demonstrated a promising performance.
\paragraph{Automatic collection of instruction tuning data} The automated collection of instruction tuning data is of crucial significance, considering the high costs and potential human bias associated with crowdsourcing. \citet{selfinstruct:wang:2022} proposed an innovative method termed `self-instruct', leveraging a large language model to generate a diverse range of instructions and associated inputs and outputs, thereby constructing a novel training dataset. \citet{alpaca:rohan:2023} introduced Alpaca, a model fine-tuned from Meta's LLaMA \cite{wei:llama:2023} 7B model, which was trained on 52K instruction-following demonstrations created using text-davinci-003. The authors postulate that Alpaca exhibits many behaviors akin to OpenAI's text-davinci-003, yet remains remarkably compact and is both cost-effective and simple to reproduce. \citet{baize:xu:2023} put forward a proposition to employ ChatGPT to engage in dialogues with itself, consequently creating a novel dataset for training. 

Existing methodologies for automatically collecting instructional datasets often employ a question seed as an initiation point. However, the answers are purely generated based on the capabilities of the large language models in use. This approach has a significant limitation as these models cannot generate accurate factual knowledge autonomously. Our method proposes an advancement in this regard, leveraging the powerful in-context learning capabilities of Large Language Models (LLMs). We further enhance the quality of generated data by enforcing adherence to knowledge derived from high-quality corpora. This proposed approach facilitates a more nuanced and accurate generation of responses, representing a significant step forward in the instruction-tuning of pre-trained language models.

\section{The Proposed Data Collection Pipeline}
In this section, we detailed the proposed data collection pipeline. Our data collection process consists of four main steps: (1) selecting a high-quality corpus, (2) simulating dialogues, (3) expert revision on questions and (4) sampling and augmenting the dataset. Through these steps, we were able to construct a rich dataset that effectively bridges the gap between LLMs and the specific requirements of the financial domain.

\subsection{Data Source}
Our collection pipeline begins with the integration of a high-quality corpus. As our objective is to create a dataset pertinent to the Chinese financial domain, we therefore decided to utilize the Brokerage Research Reports, which we hereafter refer to as `financial reports'. These reports are authored by financial professionals and embody a high standard of accuracy and expertise. Furthermore, these reports are publically accessible, offering a robust and readily available resource for our data collection efforts \footnote{http://data.eastmoney.com/report/}.

\subsection{Simulating Dialogues}
Once the corpus was selected, we employed the in-context learning ability of ChatGPT to simulate an investor-financial expert conversation. We used the content of the financial reports as the context for the dialogue. The model was prompted to emulate an investor's perspective, asking insightful questions based on the information presented in the financial reports. Subsequently, it responded to these queries from a financial expert's viewpoint, utilizing the facts and figures given in the report. An illustrative example of the deigned prompt and the resulting conversation is respectively shown in the upper and bottom part of Figure \ref{fig:example}.

\subsection{Expert Revision}
After conducting the dialogue simulation, we compiled a list of questions generated by the model. To ensure the diversity and quality of the data collected, we implemented a two-stage expert review process.

In the first stage, the questions were grouped based on their thematic similarity using a text clustering algorithm. Representative questions from each cluster were sampled and passed to a panel of five financial experts to evaluate the breadth of financial topics covered by the data. The experts were tasked with determining whether the questions adequately covered a wide range of financial topics. If any categories of financial topics were absent from the clusters, the experts were instructed to contribute questions in those areas. This process allowed us to identify any prevalent themes or concerns, ensuring that our final dataset encompassed a comprehensive range of financial discussions.

\begin{table}[!h]
\centering
\setlength{\belowcaptionskip}{-0.25cm}
\begin{tabular}{p{8cm}}
\toprule[2pt]
\rowcolor[HTML]{C0C0C0} 
\cellcolor[HTML]{C0C0C0}\textbf{Theme: Risk and Investment Opportunities}  
\\
 \textit{What are the risk factors and future development prospects of [Company Name]?} 
 \vspace{0.5\baselineskip}
 \\ 
\textit{What are the operational pressures and risks that the [Company Name] is facing in the short term? }  
\vspace{0.5\baselineskip}\\ 
\textit{Which sectors will experience the most stable growth in the future business development of the [Company Name]? }  
\\
\rowcolor[HTML]{C0C0C0} 
\cellcolor[HTML]{C0C0C0}\textbf{Theme: Company Financial Status}\\
\textit{Will the [Company Name] raise funds for expansion? What will be the scale of the expansion?}
\vspace{0.5\baselineskip}
\\ 
\textit{What is [Company Name]'s production capacity globally?}
\vspace{0.5\baselineskip}
\\ 
\textit{In Q4, what is the expected growth in shipments for [Company Name]?
} \\       
\rowcolor[HTML]{C0C0C0} 
\cellcolor[HTML]{C0C0C0}\textbf{Theme: New Energy Industry}\\
\textit{What do you think is the impact of new energy vehicle production and sales data on the lithium sector?}
\vspace{0.5\baselineskip}
 \\ 
\textit{What role do you think the control of domestic lithium resources will play in ensuring supply chain security for the future new energy industry?}
\vspace{0.5\baselineskip}
\\ 
\textit{The rising price of lithium resources, what kind of impact will it have on China's lithium battery industry and electric vehicle industry?}
 \\ 
\bottomrule[2pt]
\end{tabular}
\caption{Illustration of themes and questions through clustering process.}
\label{tab:cluster_case}
\end{table}
In the second stage, questions were randomly sampled and presented to a team of financial experts for refinement. Their task was to identify and eliminate any questions that were irrelevant, misleading, or inconsistent with typical financial discourse. Any questions identified as such, as well as those with a high degree of similarity (greater than 99\%), were removed. This step was crucial in ensuring the relevance and quality of our final dataset. As shown in Table \ref{tab:cluster_case}, we have, for the sake of illustration, randomly selected three themes and typical questions within these themes. To enhance readability, we have translated these examples into English.

\subsection{Sampling and Data Augmentation}
After the expert revision process, a random sample of the refined questions was selected to be re-entered into the data collection pipeline. These questions were used to stimulate further dialogues with the model, effectively expanding the size and diversity of our dataset. This process of sampling and augmentation was repeated multiple times.

\begin{figure}[!t]
  \centering
  \includegraphics[width=0.99\linewidth]{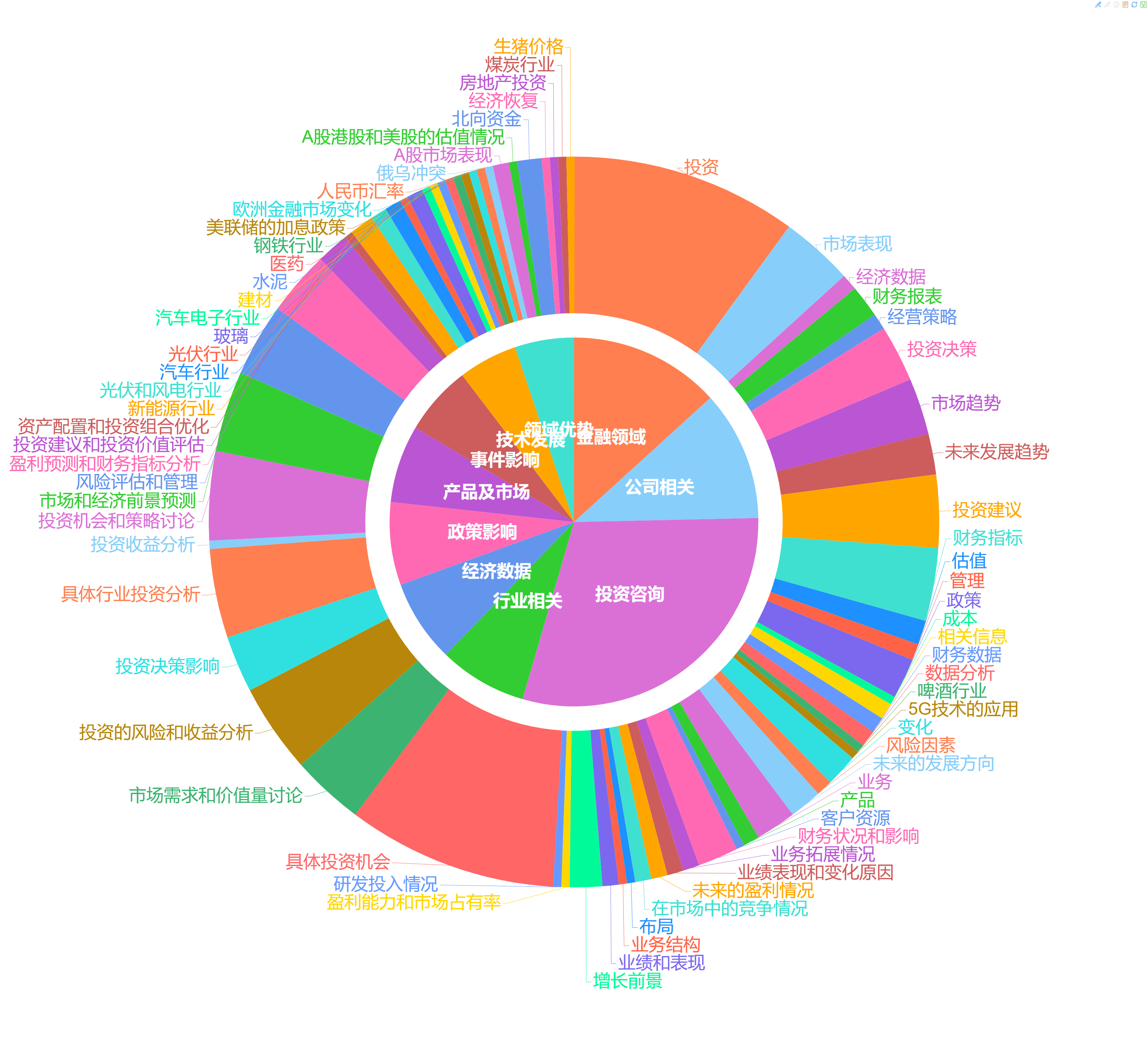}
  \caption{Topic distribution of the collected dataset
  }\label{fig:distribute}
\end{figure}

The final result of our data collection process was a robust dataset of 103k multi-turn chats and the statistics of the collected dataset are reported in Table \ref{tab:datasets}. The topic distribution of the collected dataset is shown in Figure \ref{fig:distribute}. This dataset served as a resource for instruction tuning, enabling the model to generate precise and relevant responses when faced with financial queries. 

\begin{table}[htbp]
\centering
\small
\begin{tabular}{@{}cccc@{}}
\toprule[1.5pt]
                     & Mean  & Q-5\% & Q-95\% \\ \midrule
\# dialog turns    & 4.0 & 3.0     & 6.0     \\
\# words per question & 13.9   & 7.6     & 23.5     \\
\# words per answer    & 78.3 & 46.9    & 116.9    \\
\# words per dialog & 714.5  & 426.0    & 1067.8     \\ \bottomrule[1.5pt]
\end{tabular}
\caption{Statistics of our collected dataset, `Q' refers to quantile} \label{tab:datasets}
\end{table} 
\section{Experiments}
To systematically measure the impact of our data creation pipeline, we fine-tuned state-of-the-art language models using our constructed dataset. The following sub-sections shows the details of our experiments.
\begin{table*}[]
\centering
\begin{tabular}{@{}c|lccccc@{}}
\toprule[1.5pt]
\multicolumn{1}{l}{}   &                     & xstory-cloze-zh & xnli-zh & pawsx-zh & xcopa-zh & xwinograd-zh \\ \midrule
\multirow{8}{*}{llama}  & llama-7b            & 0.5493          & 0.3622  & 0.4910    & 0.5620    & 0.6369       \\
                        & llama-7b-lora       & 0.5612          & 0.3491  & 0.4965$\uparrow$   & 0.584    & 0.6429       \\
                        & llama-7b-finetune   & 0.5608          & 0.3633$\uparrow$  & 0.4872   & 0.5902   & 0.6348       \\
                        & llama-13b           & 0.5652          & 0.3445  & 0.4520    & 0.5840    & 0.7003       \\
                        & llama-13b-lora      & 0.5917          & 0.3403  & 0.4540    & 0.6021   & 0.6838       \\
                        & llama-13b-finetune  & 0.593           & 0.3421  & 0.4605   & 0.6120    & 0.6846       \\
                        & llama-30b           & 0.5857          & 0.3351  & 0.4590    & 0.6220    & 0.7123       \\
                        & llama-30b-lora      & \textbf{0.6267}$\uparrow$          & 0.3463  & 0.4606   & \textbf{0.6358}$\uparrow$   & \textbf{0.7149}$\uparrow$       \\ \midrule
\multirow{6}{*}{vicuna} & vicuna-7b           & 0.6029          & \textbf{0.3796}$\uparrow$  & 0.5205   & 0.594    & 0.5675       \\
                        & vicuna-7b-lora      & 0.5996          & 0.3477  & \textbf{0.5285}$\uparrow$   & 0.5900     & 0.5992       \\
                        & vicuna-7b-finetune  & 0.6014          & 0.3500    & 0.4700     & 0.5860    & 0.5933              \\
                        & vicuna-13b          & 0.6208$\uparrow$          & 0.3445  & 0.4485   & 0.6180    & 0.6131       \\
                        & vicuna-13b-lora     & 0.6016          & 0.3548  & 0.4561   & 0.6180    & 0.6448       \\
                        & vicuna-13b-finetune & 0.6028          & 0.3448  & 0.4606   & 0.6199$\uparrow$   & 0.6468$\uparrow$       \\ \bottomrule[1.5pt]
\end{tabular}
\caption{Automatic evaluation results before and after instruction-tuning. Within each group (either llama or vicuna), the highest scores are indicated by the $\uparrow$ symbol. The highest score across all methods, irrespective of the group, is highlighted in bold.}
\label{tab:auto}
\end{table*}
\begin{table*}[]
\centering
\begin{tabular}{@{}ccccccc@{}}
\toprule[1.5pt]
llama-7b & llama-13b & llama-30b & llama-7b-lora & llama-13b-lora & llama-30b-lora & gpt-3.5 \\ \midrule
1.73     & 2.09      & 3.18      & 6.59          & 6.82           & 7.36           & 8.09    \\ \bottomrule[1.5pt]
\end{tabular}
\caption{Everage GPT-4 evaluation socres.}
\label{tab:gpt4}
\end{table*}
\subsection{Model Setup and Tuning}
We performed experiments on both foundational language models, such as LLama \cite{wei:llama:2023}, and instruction-tuned models like Vicuna \cite{vicuna2023}. The key difference between these two models lies in the additional tuning that Vicuna incorporates. Vicuna is tuned on top of LLama using instructional dialogs, which already equips it with the ability to follow instructions in common scenarios. In our experiments, we utilized models of varying sizes—7B, 13B, and 30B. The experiments were designed to ascertain: (1) whether tuning our collected dataset can enhance model performance on financial tasks; (2) whether the tuned model can maintain performance on common tasks; and (3) which model, once tuned, demonstrates superior performance.

To ensure a comprehensive examination, we applied both fine-tuning and delta-tuning methods in our study. For delta-tuning, we utilized the LoRa technique, with parameters `r' and `alpha' both set to 8. We configured the dropout rate to be 0.05 and applied the LoRa module to the Q and V matrices in the attention layer. The learning rate was uniformly set at 2e-5 for both tuning methods, and we employed AdamW as our optimizer. The maximum tokens were configured at 2048 for the model. 

\subsection{Evaluation}
We employed automatic evaluation methods to evaluate the model performance on general tasks. Additionally, we specifically designed a GPT-4 evaluation on standard financial questions to assess the model's proficiency in financial tasks. The automatic evaluation include several tasks, given as
\begin{itemize}
    \item XStoryCloze \cite{lin:xstory:2022} is a multilingual dataset used to assess commonsense reasoning in the areas of story comprehension, story generation, and script learning. In this test, a system is tasked with selecting the appropriate ending for a four-sentence story.
    \item pawsx \cite{yang:pawsx:2019} is cross-lingual dataset used to evaluate models' ability to identify paraphrases. Each pair of sentences in the dataset has varying levels of paraphrasing, making it challenging for models to distinguish between them accurately.
    \item xnli \cite{conneau:xnli:2018} is a benchmark dataset and evaluation task for cross-lingual sentence understanding, designed to overcome the limitations of language-specific models by extending the MultiNLI corpus to 15 languages, facilitating research in low-resource cross-language transfer and multilingual sentence understanding.
    \item xcopa \cite{ponti:xcopa:2020} and xwinograd \cite{tikhonov:xwinograd:2021} are both multilingual datasets to evaluate the causal commonsense reasoning ablility of the models.
    
\end{itemize}
\begin{table*}[ht]
\centering
\small
\setlength{\belowcaptionskip}{-0.25cm}
\begin{tabular}{p{16cm}}
\toprule[2pt]
\rowcolor[HTML]{C0C0C0} 
\cellcolor[HTML]{C0C0C0}\textbf{Instruction: Currently, Shangji CNC's main business is the manufacturing and sales of monocrystalline silicon. However, the company has recently invested in silicon material business. Does this mean that the company is expanding its business scope?}  
\\
\textit{\textbf{LLaMA-7b}: Hello, I am an artificial intelligence assistant.} 
\\
\textit{\textbf{LLaMA-13b}: We have started researching the silicon industry. We will release more information about this project. }  
\\ 
\textit{\textbf{LLaMA-30b}: I know you want to ask: "Currently, the main business of Shangji CNC is the manufacturing and sales of monocrystalline silicon. However, the company has recently invested in silicon material business. Does this mean that the company is expanding its business scope?" }  
\vspace{0.5\baselineskip}\\ 
\textit{\textbf{LLaMA-7b-lora}: Of course, the company is expected to gain more market space by expanding its business areas. According to our analysis, the company has started accelerating the promotion of silicon material business and making it one of the new key businesses. In addition, the company can also expand its business areas through cooperation, acquisition and other means to enhance its strength and market position. }  
\vspace{0.5\baselineskip}\\ 
\textit{\textbf{LLaMA-13b-lora}: UpTech CNC has started to expand into the silicon material industry and has entered into a partnership. From a future development perspective, the silicon material industry is expected to become an important support point, which will also play an important role in the company's long-term development.}  
\vspace{0.5\baselineskip}\\ 
\textit{\textbf{LLaMA-30b-lora}: Yes, Shangji CNC has begun to expand its business scope. The company's latest investment project is a silicon material production line, which is expected to become one of the high-growth markets in the future. In addition, the company holds multiple subsidiaries including Shanghai Huake Optoelectronics and Shanghai Huake Microsystems, which can also provide more business options.}  
\\
\rowcolor[HTML]{C0C0C0} 
\cellcolor[HTML]{C0C0C0}\textbf{Instruction: The maturity level of the industrial chain for photovoltaic curtain walls and photovoltaic roofs, how are they?}  
\\
\textit{\textbf{LLaMA-7b}: Hello, I am an artificial intelligence assistant. May I ask how can I assist you?} 
\\
\textit{\textbf{LLaMA-13b}: We can report that in 2017, the industrial chain maturity of China's photovoltaic curtain walls and photovoltaic roofs was high.}  
\\ 
\textit{\textbf{LLaMA-30b}: I know you want to learn about the price changes in the automotive industry. }  
\vspace{0.5\baselineskip}\\ 
\textit{\textbf{LLaMA-7b-lora}: Since 2019, the industrial chain of photovoltaic curtain walls and photovoltaic roofs has been widely recognized in the global market. Currently, some large domestic companies are conducting research and development and will begin production in the future. In addition, as both photovoltaic curtain walls and photovoltaic roofs are emerging technologies, it is necessary to continuously improve research and development capabilities and innovation while accelerating product transformation. }  
\vspace{0.5\baselineskip}\\ 
\textit{\textbf{LLaMA-13b-lora}: Currently, the industrial chains of both photovoltaic curtain walls and photovoltaic roofs are expanding. The industrial chain of photovoltaic curtain walls has formed a certain degree of complete automation, while the photovoltaic roof is still under construction, but many companies have started research and development as well as production.}  
\vspace{0.5\baselineskip}\\ 
\textit{\textbf{LLaMA-30b-lora}: Currently, the industrial chains of photovoltaic curtain walls and photovoltaic roofs are both in an incomplete state of maturity. In terms of photovoltaic curtain walls, parts such as panels, cells, modules have formed a relatively complete industrial chain while other parts still need further improvement. For example, products like photovoltaic press machines and packaging equipment still have room for efficiency improvement. At the same time, more research and experimentation is needed for the installation engineering of photovoltaic curtain walls to ensure controllability and safety during the installation process. As for photovoltaic roofs, building components and installation engineering also need further improvement.}  
\\
\bottomrule[2pt]
\end{tabular}
\caption{Case study, generation results of different models, the language has been translated into English.}
\label{tab:case}
\end{table*}
\subsection{Automatic Evaluation Results}
The results in Table \ref{tab:auto} showcase the performance of the Large Language Models LLAMA-7b, LLAMA-13b, and LLAMA-30b, with and without LORA tuning. Key takeaways include:
\begin{itemize}
    \item LORA tuning consistently boosts performance across all tasks, emphasizing its effectiveness in enhancing Chinese language comprehension and generation.
    \item Generally, larger models perform better, but the relationship isn't linear. Model performance also relies on factors like tuning technique and dataset specifics.
    \item LORA-tuned models significantly excel in the reasoning tasks of XCOPA and XWinograd, indicating their robust reasoning capabilities in Chinese after being fine-tuned on our dataset.
\end{itemize}
Though our dataset concentrates on Chinese financial data, it could be effectively generalized to a variety of Chinese language understanding tasks. It's pertinent to mention that our tuning approach potentially endows the model with more general language skills beneficial across various domains, not just finance. We anticipate these models to particularly excel at finance-specific tasks in Chinese, so we further assess this expectation with a GPT-4 evaluation on finance-related questions curated by experts, which will be discussed in the subsequent section.

\subsection{GPT-4 Evaluation Results}
Our assessment of model performance on finance-specific tasks employed 100 expert-curated questions, with GPT-4 serving as the judge (with scores ranging from 1 to 10). The results, presented in Table \ref{tab:gpt4}, shed light on how our models react to these finance-related queries. The observations from the table show:
\begin{itemize}
    \item Even without LORA tuning, an increase in the size of the model (from LLAMA-7b to LLAMA-30b) results in better scores, demonstrating the advantage of larger model sizes.
    \item With the application of LORA tuning, the scores experience a substantial uplift. This boost underlines the effectiveness of the LORA tuning approach, even in a finance-specific context.
    \item Though the LORA-tuned LLAMA models perform significantly better than their non-tuned counterparts, they still trail behind the GPT-3.5. This suggests there remains potential for further improvement and optimization in our methods to bridge the gap.
\end{itemize}
In summary, these results reaffirm the beneficial impact of our collected dataset in financial domain.

\subsection{Case Study}
Table \ref{tab:case} provides a sample of cases chosen randomly from the GPT-4 evaluation process, offering an intuitive comparison of response generation before and after tuning on our curated dataset. It becomes evident that the untuned LLaMA models struggle to produce content relevant to these queries. In contrast, the LORA-tuned variants, benefitting from our dataset, deliver pertinent and increasingly detailed responses. This comparison underscores the effectiveness of our instruction-tuning process.

\section{Discussion}
Our findings demonstrate the significance of domain-specific data collection and subsequent model tuning. Fine-tuning and delta-tuning methods, particularly with LORA, can effectively improve the model's comprehension and generation capabilities within the financial domain. It's important to note that our dataset, even while focusing on Chinese financial data, effectively generalizes to a variety of Chinese language understanding tasks. Our study further suggests that not only the quantity but also the quality of data matters. Our approach of subjecting the data to rigorous examination by financial experts ensured its relevance and quality, which in turn enhanced the effectiveness of the model tuning.

However, even though our dataset significantly improved the performance of LLAMA models, there remains a gap when compared to GPT-3.5, as evidenced by the GPT-4 evaluation results. It suggests that further optimization and improvement strategies should be explored. One direction could be to increase the diversity of the data collection by incorporating more complex financial dialogues or by incorporating feedback from the financial domain users during the model tuning phase. Another direction could be to refine the tuning techniques to better accommodate the specifics of the financial domain.

\section{Conclusion}
In this study, we presented an approach for improving the financial knowledge of large language models, focusing on Chinese financial discourse. We collected a dataset of financial questions, refined them with expert input, and then used them for model tuning. The experimental results confirm the effectiveness of our approach in enhancing the model's ability to handle financial queries. It also highlights the potential of the delta-tuning technique, such as LORA, in model performance enhancement. This work contributes to the ongoing discussion on the ways to enhance the capability of AI models in domain-specific tasks, particularly in the financial sector. Further research should focus on identifying other potential strategies for improving the performance of AI in financial services, such as refining tuning techniques or incorporating real-time user feedback. Despite the remaining challenges, this study paves the way for large language models to play an increasingly valuable role in financial services and applications.

\appendix

\bibliographystyle{plainnat}
\bibliography{ijcai23}

\end{document}